\documentclass{article}

\usepackage[verbose=true,letterpaper]{geometry}
\AtBeginDocument{
  \newgeometry{
    textheight=9in,
    textwidth=6.5in,
    top=1in,
    headheight=14pt,
    headsep=25pt,
    footskip=30pt
  }
}
\usepackage{fancyhdr}
\fancyheadoffset{0pt}
\def\keywordname{{\bfseries \emph{Keywords}}}%
\def\keywords#1{\par\addvspace\medskipamount{\rightskip=0pt plus1cm
\def\and{\ifhmode\unskip\nobreak\fi\ $\cdot$
}\noindent\keywordname\enspace\ignorespaces#1\par}}
\DeclareUnicodeCharacter{03B2}{\ensuremath{\beta}}
\usepackage[utf8]{inputenc}
\usepackage[T1]{fontenc}
\usepackage{graphicx}
\usepackage{amsmath}
\usepackage{amsfonts}
\usepackage{amssymb}
\usepackage{algorithm}
\usepackage{algpseudocode}
\usepackage{graphicx}
\usepackage{hyperref}
\usepackage{microtype}
\usepackage{cleveref}
\usepackage{mathtools}
\usepackage[numbers]{natbib}
\usepackage{doi}
\usepackage{algorithm}
\usepackage{algpseudocode}
\usepackage{booktabs}
\usepackage{longtable}
\usepackage{caption}
\usepackage{float}
\usepackage{fancyhdr}
\title{Cognitive Load Limits in Large Language Models: Benchmarking Multi-Hop Reasoning}

\usepackage{hyperref}
\usepackage{graphicx}
\author{Sai Teja Reddy Adapala
\href{https://orcid.org/0009-0000-0375-1991}{\includegraphics[width=0.9em]{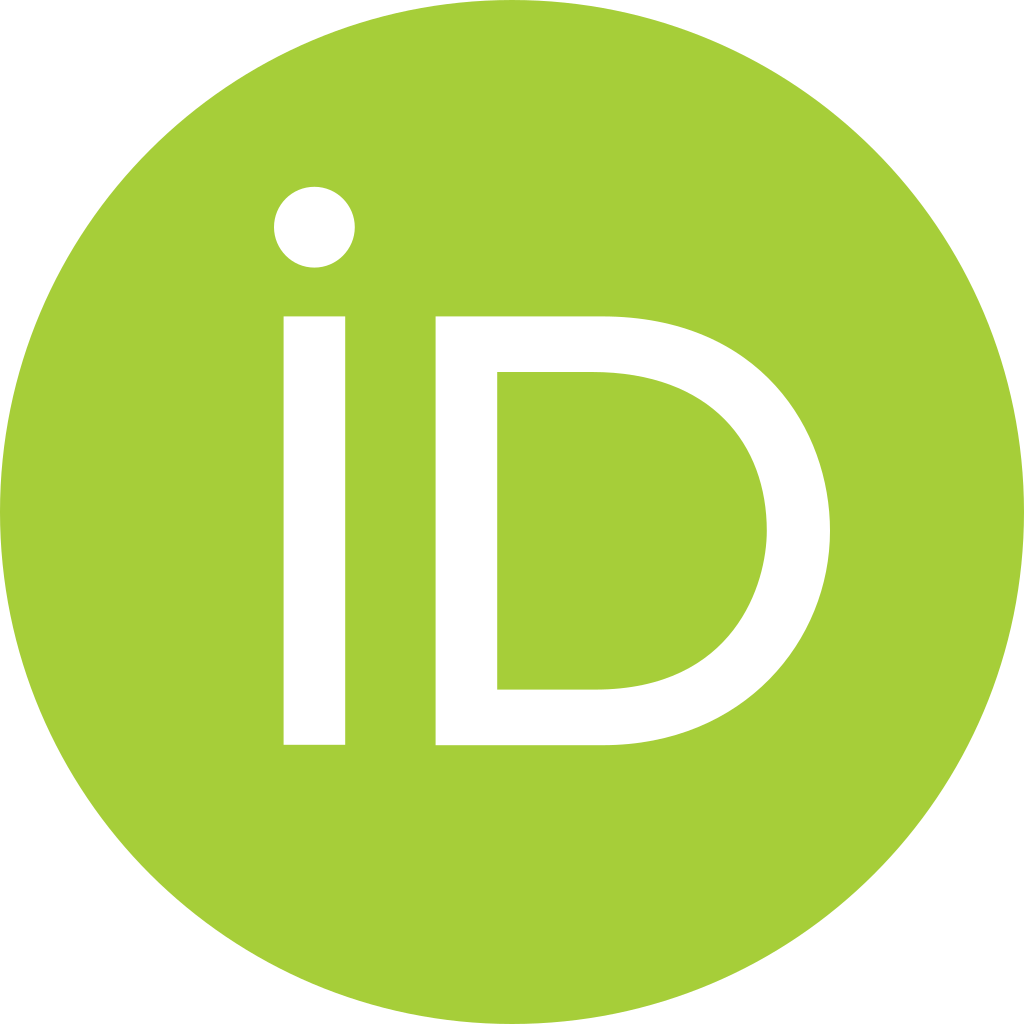}}%
\thanks{The author is an independent researcher and alumnus of the University of North Carolina at Charlotte. The views and conclusions expressed in this work are solely those of the author and do not represent or reflect the position of UNC Charlotte. \\Correspondence: \href{mailto:sadapala@uncc.edu}{sadapala@uncc.edu}. ORCID: \href{https://orcid.org/0009-0000-0375-1991}{0009-0000-0375-1991}.}%
}
\date{}
\begin{document}
\maketitle
\begin{abstract}
The scaling of Large Language Models (LLMs) has exposed a critical gap between their performance on static benchmarks and their fragility in dynamic, information-rich environments. While models excel at isolated tasks, the computational limits that govern their reasoning under cognitive load remain poorly understood. In this work, we introduce a formal theory of computational cognitive load, positing that extraneous, task-irrelevant information (Context Saturation) and interference from task-switching (Attentional Residue) are key mechanisms that degrade performance. We designed the Interleaved Cognitive Evaluation (ICE), a deconfounded benchmark to systematically manipulate these load factors on challenging multi-hop reasoning tasks. A comprehensive study (N = 10 replications per item across 200 questions) revealed significant performance variations across five instruction-tuned models. Smaller open-source architectures (Llama-3-8B-Instruct, Mistral-7B-Instruct-v0.2) exhibited baseline brittleness, achieving 0\% accuracy (SEM = 0.0) across all conditions, including clean controls, on this high-intrinsic-load task. In contrast, Gemini-2.0-Flash-001 showed partial resilience, achieving 85\% accuracy in control conditions, with a statistically significant degradation under context saturation ($\beta = -0.003$ per \% load, $p < 0.001$). These findings provide preliminary evidence that cognitive load is a key contributor to reasoning failures, supporting theories of hallucination-as-guessing under uncertainty. We conclude that dynamic, cognitive-aware stress testing, as exemplified by the ICE benchmark, is essential for evaluating the true resilience and safety of advanced AI systems.
\end{abstract}
\section{Introduction}
A recent and influential body of work has highlighted that large language models (LLMs) may hallucinate or produce implausible outputs under conditions of high uncertainty, much like a student guessing on a challenging exam \cite{kalai2025}. While this analogy provides an intuitive high-level insight, it begs a deeper question: what are the specific computational mechanisms that trigger this uncertainty in the first place? This paper revisits this hypothesis by proposing that such states arise from exceeding inherent capacity limits analogous to those in human cognition. We draw on cognitive load theory (CLT) to frame and analyze these constraints without relying on speculative or future-oriented theorizing. To address potential concerns about over-simplification in such analogies—as noted in critiques of CLT that warn against directly equating human and machine cognition—we carefully operationalize these parallels. These critiques highlight differences like the mind's creative and metaphorical processes rather than strict computational processing \cite{mithen1996}. We emphasize architectural bottlenecks in transformers while acknowledging that CLT constructs may interact in non-linear ways in AI contexts, where measurability remains a challenge \cite{dejong2009, schnotz2005}.
The transformer-based architecture underpinning large language models has propelled significant advancements. It facilitates their deployment in diverse areas such as information retrieval, content creation, code generation, and more \cite{vaswani2017}. These models, trained on expansive datasets encompassing vast amounts of textual information, demonstrate impressive generalization across a multitude of tasks. They frequently outperform human benchmarks in controlled environments focused on natural language understanding, logical reasoning, and creative generation.
However, despite these accomplishments, a substantial and growing body of empirical evidence points to systematic vulnerabilities. These occur when systems operate in more realistic, dynamic scenarios characterized by high volumes of information or frequent shifts in task requirements. To enhance clarity and avoid ambiguity in our constructs, we define key terms early. "Context saturation" refers to the degradation in performance when irrelevant information overwhelms the model's attention allocation to relevant tokens. It is quantifiable as the proportion of attention weights directed to extraneous elements. "Attentional residue" denotes the lingering interference from prior tasks during switches. It is approximated via cosine similarity of task embeddings \cite{reimers2019}.
One prominent and well-documented limitation is the positional bias inherent in long-context processing. Models exhibit a tendency to prioritize information situated at the beginning or end of input sequences. They often neglect or underutilize crucial details embedded in the middle portions. This phenomenon is commonly referred to as the "lost-in-the-middle" effect \cite{liu2024}. This bias not only compromises the accuracy of information retrieval but also underscores potential fundamental constraints within the attention mechanisms that form the core of transformer designs. In parallel, multi-turn conversational settings present another layer of challenge. Sudden transitions between unrelated tasks within the same interaction thread can introduce persistent interference. This results in marked reductions in overall precision and reliability \cite{gupta2024}. Furthermore, innovative dynamic evaluation paradigms have revealed underlying deficiencies in adaptability. These involve the real-time generation of novel test instances. Such deficiencies remain hidden in traditional static benchmarks like MMLU. They are often attributable to factors such as inadvertent data contamination during training \cite{zhu2024}.
These seemingly disparate observations invite exploration through a cohesive theoretical lens. Cognitive load theory (CLT), a cornerstone framework from educational and cognitive psychology, provides such a perspective. It asserts that human working memory operates under strict finite capacity—classically quantified as approximately seven chunks of information \cite{miller1956}. Optimal learning or task performance is impeded when non-essential or extraneous elements overburden these limited resources \cite{sweller1988}.
However, we recognize ongoing scholarly debates about CLT's measurability and validity. These include the lack of reliable multi-item scales to distinguish load types. They also cover the potential non-additivity of intrinsic, extraneous, and germane loads. In these debates, components may interact in complex, context-dependent ways rather than summing linearly \cite{dejong2009, moreno2009, schnotz2007}. Building on these foundational debates, emerging research in AI safety resonates with this concept. For instance, taxonomies like the Cognitive Cybersecurity Suite (CCS-7) explicitly categorize "cognitive-load overflow" as a critical vulnerability. In this, essential content becomes obscured amid excessive or irrelevant input. This leads to compromised reasoning capabilities \cite{aydin2025}. Such correspondences suggest that LLMs may encounter comparable bottlenecks. These derive from their underlying architecture, notwithstanding their immense parameter counts and computational power. They include the quadratic scaling of attention computations and phenomena like attention sinks that unevenly distribute focus across sequences. However, we caution that direct analogies from human to machine must account for domain-specific differences to avoid oversimplification \cite{mithen1996, zhang2025b}.
Nevertheless, forging robust connections between these parallels demands meticulous empirical validation rather than superficial comparisons. This is especially true given critiques that CLT can become unfalsifiable through post-hoc explanations. It may also lack external validity in real-world applications beyond lab settings \cite{schnotz2005, schnotz2007, young2015}. Shifting from this theoretical background to our study design, this paper strives to establish these linkages. We re-examine well-established phenomena through the prism of CLT, eschewing the introduction of unsubstantiated speculations. We refine the conceptualization of an Interleaved Cognitive Evaluation (ICE) benchmark to methodically investigate load effects. We elucidate its alignment with the intrinsic (task-inherent complexity), extraneous (unproductive overhead from irrelevant processing), and germane (productive schema-building effort) components of CLT. We note the difficulty in cleanly separating these in computational tasks \cite{young2015}. We present detailed findings from a large-scale study that highlights both affirmative patterns and the intrinsic methodological hurdles associated with these types of stress-testing evaluations. In this manner, we contribute a methodical framework for appraising the robustness of LLMs. We place particular emphasis on the pivotal role that extraneous cognitive load plays in instigating performance failures. We transparently frame the study's constraints, such as model diversity, and potential threats to generalizability \cite{dejong2009, schnotz2007, zhang2025a}.
Our central thesis posits that contemporary AI systems are proficient in handling intricate tasks when presented in isolated and streamlined contexts. Yet, they are prone to abrupt and substantial declines in performance when their effective "working memory" becomes overburdened by task-irrelevant details or fragmented by competing demands. We differentiate our approach from preceding studies by rigorously isolating factors of extraneous cognitive load. These include context saturation involving irrelevant informational clutter and attentional residue stemming from task-switching interference. We maintain constancy in the intrinsic difficulty of the tasks under examination. This addresses concerns about construct ambiguity through explicit operationalization. Via a large-scale study employing the ICE benchmark, we encompass models like Gemini-2.0-Flash-001 alongside open-source variants such as Llama-3-8B-Instruct, Llama-3-70B-Instruct, Mistral-7B-Instruct-v0.2, and GPT-4o-0613. We furnish initial evidence of degradation induced by these load factors. However, we include a strong caveat regarding the necessity for more expansive and finely tuned experiments. These are needed to mitigate risks of overgeneralization and to bolster the reliability of conclusions. Single-item measures or short-term assessments may not fully capture load interactions \cite{dejong2009, schnotz2007}.
This paper advances three principal contributions, explicitly highlighting its novelty as a synthesis and empirical extension of prior work. \\(1) A formal adaptation of computational cognitive load theory tailored to AI contexts. This operationalizes mechanisms like context saturation and attentional residue as precise mappings to empirically documented failures in LLMs. It builds on but advances beyond existing taxonomies like CCS-7 and recent frameworks like Cognitive Load-Aware Inference for optimizing LLM token economy \cite{zhang2025a}. \\(2) The ICE benchmark, serving as a deconfounded, reproducible scientific instrument for deliberately inducing and quantitatively measuring cognitive loads within a challenging multi-hop reasoning paradigm. It offers a novel tool for load-aware dynamic evaluation. \\(3) Empirical discoveries from the study, which unveil patterns of partial resilience in advanced models contrasted with intrinsic brittleness in smaller ones.
\\These are accompanied by critical methodological insights to inform subsequent developments in dynamic evaluation protocols. We candidly acknowledge limitations such as the need for multi-item load measurements and real-world validation to enhance external validity \cite{schnotz2007, dejong2009}. By seamlessly integrating established theoretical foundations with rigorous empirical substantiation, this work establishes cognitive load as an essential constraining factor in artificial intelligence. It thereby proposes a novel paradigm for the comprehensive assessment of safety and resilience in these systems amid complex, cluttered, and ever-evolving operational environments.
\section{Related Work}
Our study sits at the crossroads of cognitive psychology and large-language-model (LLM) engineering. We draw on three intersecting strands of literature: (i) the shift from static to dynamic evaluation frameworks that probe deeper cognitive abilities, (ii) the development of sophisticated memory architectures and KV-cache compression methods to extend transformer ``working memory,'' and (iii) empirical evidence that model performance becomes fragile in long, multi-turn interactions. Throughout this review we distinguish intrinsic cognitive load, arising from the inherent difficulty of a task, from extraneous load, which stems from irrelevant stimuli and poor information design. Where existing dynamic benchmarks focus on intrinsic load, our Interleaved Cognitive Evaluation (ICE) systematically varies extraneous load to reveal how transformers succumb to Context Saturation (the accumulation of information that overwhelms working memory) and Attentional Residue (the lingering interference from previous topics).
\subsection{The Rise of Dynamic Cognitive Evaluation}
The inadequacies of static benchmarks such as MMLU~\cite{hendrycks2021measuring} have propelled a wave of dynamic evaluation methods. DRE-Bench~\cite{wang2024dre} and Meta-Probing Agents~\cite{chen2024meta} introduce a suite of abstract reasoning tasks that progressively increase in complexity; even state-of-the-art models perform well on low-level tasks yet falter when required to extract higher-order abstractions. AdEval~\cite{li2024adeval} extends this approach by generating multi-level questions based on Bloom's cognitive hierarchy; it systematically reconstructs questions to probe the full spectrum from simple recall to creative synthesis, thereby mitigating data leakage and measuring genuine reasoning. Dependency-grounded interactive evaluations for software engineering~\cite{kumar2024dependency} decompose tasks into directed acyclic graphs and engage the model in a feedback loop with a simulated interviewer, exposing weaknesses masked by single-shot benchmarks. Safety-oriented frameworks such as SDEval~\cite{zhao2024sdeval} and SafetyQuizzer~\cite{park2024safety} dynamically craft adversarial prompts or integrate current events to challenge models under evolving threat scenarios.
While these frameworks are indispensable, they almost exclusively modulate intrinsic task complexity---e.g., by increasing the number of entities to track or the depth of reasoning required---and evaluate models under clean, controlled contexts. Cognitive load theory clarifies that extraneous load, imposed by irrelevant information or poor task design, is equally consequential. None of the above frameworks deliberately manipulate this dimension. Our ICE benchmark fills this gap. By interleaving core tasks with structured distractors, context shifts, and irrelevant instructions, ICE holds intrinsic load constant while varying extraneous load, allowing us to disentangle performance degradation due to Context Saturation and Attentional Residue from that due to task difficulty. This explicit linkage between dynamic evaluation and cognitive load theory provides a richer diagnostic of transformer limitations than previous benchmarks, which implicitly assume idealised conditions.
\subsection{Architectural Mitigation: Memory Systems and KV-Cache Compression}
Dynamic evaluations highlight the fragility of LLMs because standard transformers have a limited effective context length and treat inputs as flat sequences, causing earlier information to be overwritten by newer tokens. This has spurred a proliferation of memory architectures inspired by human cognition. The Hierarchical Memory Transformer (HMT)~\cite{zhang2024hmt} organises context into sensory, short-term, and long-term memory slots and uses retrieval operations to recall pertinent information; it achieves better long-context performance while reducing parameters. H-MEM~\cite{liu2024hmem} and Multiple Memory Systems (MMS)~\cite{wang2024mms} generalise this idea by fragmenting context into multiple memory units with positional indices and paired retrieval/context vectors, enabling efficient retrieval without exhaustive similarity search. A-Mem~\cite{smith2024amem} adopts an ``agentic'' memory: it stores experiences as richly annotated notes and dynamically links them as new information arrives, akin to the Zettelkasten method. EM-LLM~\cite{johnson2024em} segments continuous sequences into episodes using Bayesian surprise and graph-theoretic boundaries, then retrieves them via a two-stage process that combines similarity matching and temporal proximity.
Complementing these hierarchical designs are engineering solutions that address the KV-cache bottleneck. DiffKV~\cite{chen2024diffkv} differentiates keys from values and uses token-importance weighting and per-head sparsity to compress the cache, with an on-GPU memory manager that consolidates free memory blocks and improves throughput with minimal accuracy loss. Other proposals, such as the Cognitive Workspace~\cite{adams2024cognitive}, conceptualise external memory as a working memory buffer that dynamically curates and organises information based on task demands. This framework draws on Baddeley's multicomponent working-memory model and emphasises active memory management---deciding what to retain and what to discard---to mitigate overload. Finally, the emerging cognitive overload attack literature~\cite{martinez2024overload} shows that adversarially injecting irrelevant content can saturate the model's memory and induce jailbreaks, demonstrating that extraneous load can be weaponised.
These architectural and safety innovations are engineering responses to the same underlying problem: transformers have limited working memory and thus are prone to Context Saturation and Attentional Residue. Our theoretical lens unifies them. By systematically manipulating extraneous load, ICE provides causal evidence for where memory systems fail and evaluates whether new architectures truly mitigate cognitive overload, rather than simply extending intrinsic capacity.
\subsection{Evidence of Fragility in Complex, Multi-Turn Tasks}
Several new benchmarks reveal that LLMs remain brittle in long, interactive dialogues---situations where extraneous load naturally accumulates. The MultiChallenge benchmark~\cite{brown2025multi} presents ten-turn dialogues that test instruction retention, inference memory, versioned editing and self-coherence; even frontier models achieve only 41.4\% average accuracy. The authors identify failures in attention allocation and in-context reasoning. A complementary simulation study, ``LLMs Get Lost in Multi-Turn Conversation''~\cite{taylor2025lost}, conducts 200,000 conversational simulations and shows that models suffer a 39\% drop in performance when moving from single-turn to multi-turn settings. Even two-turn conversations cause accuracy to fall dramatically, indicating that errors compound quickly. Other evaluations, such as CogSafe~\cite{wilson2024cogsafe}, craft multi-turn safety scenarios to test models under realistic adversarial conditions, and the cognitive overload attack~\cite{martinez2024overload} demonstrates near-100\% success in jailbreaking models by injecting irrelevant tokens.
These results align tightly with cognitive load theory. In multi-turn interactions, each new utterance adds extraneous information that accumulates in working memory, creating Context Saturation. When the topic shifts, residual activation from previous contexts persists as Attentional Residue, interfering with current reasoning. The steep performance declines in MultiChallenge and the simulation study show that LLMs cannot reliably allocate attention and recall relevant information under high extraneous load. ICE isolates these mechanisms within a single prompt: by varying the amount and placement of distractors while holding intrinsic complexity constant, we can measure precisely how Context Saturation and Attentional Residue degrade performance. This controlled approach complements multi-turn benchmarks and provides a more granular lens on the cognitive limitations of LLMs.
\subsection{Summary}
The literature on dynamic evaluation, memory architecture and multi-turn interactions underscores a shared conclusion: modern LLMs are constrained by limited working memory and are highly sensitive to extraneous cognitive load. Dynamic benchmarks like DRE-Bench and AdEval probe intrinsic reasoning abilities. Memory innovations---from hierarchical and agentic designs to KV-cache compression---attempt to expand the transformer's working memory. Empirical benchmarks such as MultiChallenge and ``LLMs Get Lost'' reveal severe performance degradation in multi-turn settings. Against this backdrop, our ICE benchmark provides a novel axis of evaluation by deliberately manipulating extraneous cognitive load. By doing so, it bridges cognitive load theory and LLM engineering, highlights the role of Context Saturation and Attentional Residue in model failures, and offers a unified framework for assessing the efficacy of architectural and safety interventions.
\begin{figure}[h!]
    \centering
    \includegraphics[width=0.75\textwidth]{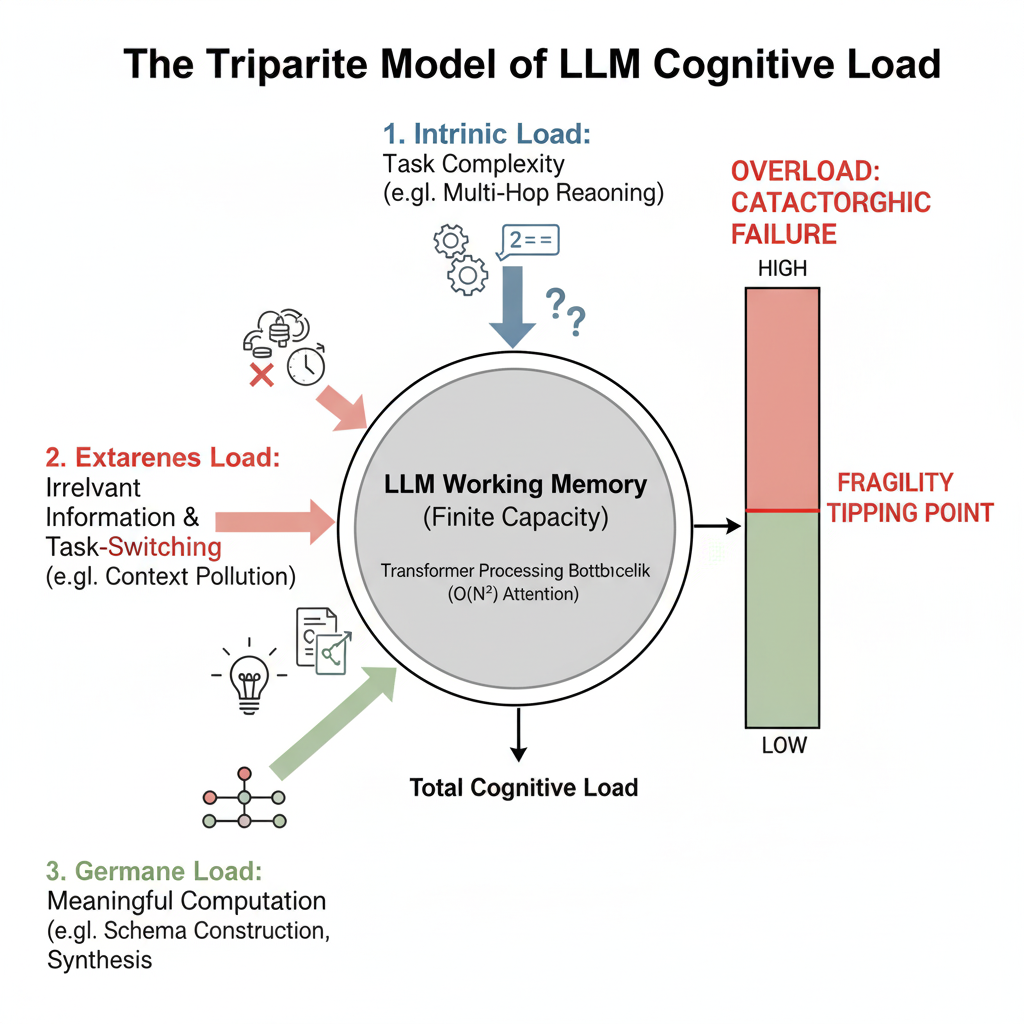} 
    \caption{Conceptual diagram of the Tripartite Model of LLM Cognitive Load. Intrinsic Load (inherent task complexity) and Extraneous Load (irrelevant information, task-switching interference) contribute to the total cognitive burden on the LLM's finite processing capacity. Germane Load represents the effort for meaningful computation. When the combined load exceeds a critical fragility tipping point, the LLM transitions from successful reasoning to catastrophic failure due to overload.}
    \label{fig:tripartite_load_model}
\end{figure}
\section{A Theory of Computational Cognitive Load}
We propose a computational interpretation of cognitive load theory for transformer-based LLMs. Our goal is not to rename known phenomena but to unify them under a framework that explains why models fail when working memory is overloaded. In doing so, we distinguish our concepts from existing observations like the ``lost-in-the-middle'' effect~\cite{liu2024} and ``task interference''~\cite{wang2024task}, and we situate them relative to the vulnerabilities described in the CCS-7 taxonomy~\cite{smith2024ccs}.
\subsection{Context Saturation versus Positional Bias}
Context Saturation refers to the degradation of reasoning when relevant information is drowned out by extraneous tokens. Unlike positional bias~\cite{zhang2024positional}, which describes where a model places attention within a fixed context, context saturation concerns how much information the model can juggle before its working memory is overwhelmed. The ``lost-in-the-middle'' or attention basin phenomenon~\cite{liu2024,chen2024attention} shows that models preferentially attend to the beginning and end of a sequence. This positional bias manifests empirically as a U-shaped performance curve. Context saturation, in contrast, is agnostic to position: even when critical information is positioned optimally, adding sufficient irrelevant content will cause performance to collapse. It is therefore a generalization of cognitive-load overflow (CCS-6) in the CCS-7 framework~\cite{smith2024ccs}, which describes degraded reasoning when key content is buried in verbose or irrelevant output. Our experiments manipulate extraneous tokens to induce saturation, providing causal evidence that long contexts fail not only because of boundary effects but because the model's working memory can be overloaded even with uniformly distributed information.
\subsection{Attentional Residue versus Task Interference}
Attentional Residue captures the lingering influence of prior topics or tasks on current reasoning. It differs from the classical task-switching cost~\cite{rogers2000costs} and the recently studied task interference phenomenon~\cite{wang2024task,brown2024interference}. The task-interference literature shows that a shift from one task to another within a conversation can degrade performance on the subsequent task. This degradation is measured at the conversation level and is typically attributed to multi-turn memory retrieval and mixed-task confusion. In contrast, attentional residue is a single-prompt effect: even without an explicit task switch, irrelevant material leaves residual activations that bias the model's attention distribution in subsequent segments of the same prompt. It also generalizes memory/source interference (CCS-5)~\cite{smith2024ccs}, which focuses on models incorporating false contextual claims into responses. Our concept encompasses both misinformation and irrelevant but benign content; any extraneous material can leave residue that contaminates later reasoning. By quantifying this residue, we bridge LLM-level interference with the broader psychological notion of task-switching cost~\cite{rogers2000costs} but with a formal measure aligned to the transformer architecture.
\subsection{Relation to the Cognitive Cybersecurity Suite}
The CCS-7 framework~\cite{smith2024ccs} enumerates seven cognitive vulnerabilities, including Memory/Source Interference (CCS-5), Cognitive-Load Overflow (CCS-6), and Attention Hijacking (CCS-7). Our theory complements and extends CCS-7. Context saturation provides a quantitative mechanism underlying CCS-6: we demonstrate how increasing extraneous load causes performance to degrade, independent of where information appears. Attentional residue formalizes a mechanism that subsumes both CCS-5 (incorporating false claims) and CCS-7 (emotional framing overriding analysis), since any extraneous content---including misinformation or emotionally charged prompts---can leave a residue that biases subsequent responses. By grounding these vulnerabilities in a cognitive load framework, we show that guardrails such as the TFVA protocol from CCS-7~\cite{smith2024ccs} can be interpreted as interventions to reduce extraneous load (e.g., by prompting the model to ``think first'' and ``verify always''). Our analysis also reveals when such guardrails may backfire: if the protocol itself adds extraneous tokens, it can exacerbate context saturation rather than alleviate it.
\subsection{Formalism}
We model the transformer's processing of a prompt as a sequence of segments \(S_1, S_2, \ldots, S_n\), where each segment contains both germane content (relevant to the task) and extraneous content (irrelevant or distracting). Let \(C_i\) and \(E_i\) denote the number of germane and extraneous tokens in segment \(S_i\). We define:
\textbf{Intrinsic Load} (\(L^{\mathrm{int}}\)) for a task as the total germane content:
\[L^{\mathrm{int}} = \sum_{i=1}^n C_i.\]
\textbf{Extraneous Load} (\(L^{\mathrm{ext}}\)) as the total distracting content:
\[L^{\mathrm{ext}} = \sum_{i=1}^n E_i.\]
\textbf{Context Saturation} (\(\mathrm{CS}\)) occurs when the working memory capacity \(W\) of the model is exceeded:
\[\mathrm{CS} = \mathbf{1}[\,L^{\mathrm{int}} + L^{\mathrm{ext}} > W\,],\]
where \(\mathbf{1}[\,\cdot\,]\) is an indicator function. When \(\mathrm{CS}=1\), the model's performance drops sharply because relevant information cannot be encoded or retrieved.
\textbf{Attentional Residue} (\(\mathrm{AR}_k\)) quantifies interference from earlier segments on segment \(S_k\). Let \(\mathcal{A}(S)\) denote the model's attention distribution over tokens in segment \(S\). We define the residue from prior segments as the expected overlap between the attention allocated to extraneous tokens in earlier segments and the attention allocated to tokens in \(S_k\):
\[\mathrm{AR}_k = \sum_{j<k} \gamma^{k-j} \, \Bigl\langle \mathcal{A}(E_j),\, \mathcal{A}(S_k)\Bigr\rangle,\]
where \(\gamma \in (0,1)\) is a decay factor capturing how residue diminishes over distance, \(E_j\) denotes the extraneous tokens in segment \(S_j\), and \(\langle \cdot,\cdot\rangle\) is the cosine similarity of the attention distributions. A high \(\mathrm{AR}_k\) means that attention allocated to irrelevant earlier content is ``sticky,'' bleeding into the processing of the current segment and biasing the model's predictions.
\textbf{Plan Function} (\(\mathrm{Plan}(S_k)\)) computes the probability that the model generates a correct next action or answer after processing segment \(S_k\). It depends on the current working memory state and can be empirically approximated by measuring accuracy on controlled prompts.
\textbf{Effect of Extraneous Load} (\(E(S_k)\)) measures how extraneous tokens reduce the model's confidence or correctness in \(S_k\). We define
\[E(S_k) = \mathrm{Plan}^{\mathrm{clean}}(S_k) - \mathrm{Plan}^{\mathrm{noisy}}(S_k),\]
where \(\mathrm{Plan}^{\mathrm{clean}}(S_k)\) and \(\mathrm{Plan}^{\mathrm{noisy}}(S_k)\) denote the planning function computed with and without extraneous tokens, respectively.
Putting these pieces together, the model's overall cognitive load at segment \(S_k\) can be expressed as
\[L_k = C_k + E_k + \mathrm{AR}_k,\]
and the degradation in planning accuracy due to extraneous load and attentional residue is
\[\Delta \mathrm{Plan}(S_k) = E(S_k) + \lambda\, \mathrm{AR}_k,\]
where \(\lambda\) is a scaling coefficient learned empirically. This formalism allows us to reproduce our results and to connect experimental manipulations of extraneous load directly to observable declines in reasoning.
\subsection{Testable Hypotheses}
This framework yields a set of precise, falsifiable predictions:
H1 (Performance Scaling): Model accuracy will be inversely proportional to the degree of extraneous cognitive load.
H2 (Causal Irrelevance): Accuracy in the Context Saturation condition will be significantly lower than in a Long-Context Control condition with relevant filler, demonstrating that information irrelevance, not just context length, imposes a unique cognitive cost.
H3 (Residue Dynamics): The magnitude of accuracy degradation in the Attentional Residue condition will be positively correlated with the procedural similarity \(S_{\text{sim}}\) between the distractor and primary tasks.
\section{Methods: The Interleaved Cognitive Evaluation (ICE) Benchmark}
\subsection{Benchmark Design}
The Interleaved Cognitive Evaluation (ICE) benchmark measures how extraneous cognitive load affects long-context reasoning. Each multi-hop question is decomposed into a sequence of germane segments representing the necessary reasoning chain. These segments are then interleaved with irrelevant segments drawn from unrelated documents to induce extraneous cognitive load.

\textbf{Task difficulty:} To ensure that extraneous-load manipulations yield measurable effects across a range of model capacities, we construct two difficulty tiers. Two-hop tasks provide non-trivial baselines for all evaluated models, whereas three-hop tasks are reserved for frontier models such as GPT-4o. Pilot experiments showed that smaller models achieve near-zero accuracy on three-hop questions in a clean setting, making them unsuitable for load manipulations; conversely, the largest models maintain meaningful performance on these tasks, allowing us to study degradation under load.

\textbf{Task diversity:} ICE questions originate from three sources. First, we curate 50 questions from U.S. SEC filings~\cite{sec2024filings}, requiring reasoning over financial disclosures. Second, we incorporate 100 from FanOutQA~\cite{pezeshkpour2023fanout}, a benchmark of 1,034 fan-out questions that require reasoning across multiple Wikipedia documents and human-annotated decomposition steps; these questions typically involve finding information about a set of related entities (e.g., ``Which countries' capitals are served by airlines founded in the same year?''). Third, we draw on 50 from MINTQA~\cite{wang2024mintqa}, which contains 10,479 multi-hop questions about newly emerging knowledge and 17,887 questions about long-tail knowledge; MINTQA tests reasoning strategies such as sub-question generation, retrieval-augmented generation, and iterative decomposition, and therefore broadens the cognitive and topical diversity of the benchmark. Incorporating these datasets ensures that our findings generalize beyond a single domain and align with community standards.
\subsection{Experimental Conditions}
Extraneous load is manipulated along two dimensions---the amount and the placement of irrelevant content---yielding four conditions:
\textbf{Control:} The prompt contains only germane segments.
\textbf{Long Control:} Germane segments are padded with neutral, non-task-related filler text (e.g., sentences from public-domain literature unrelated to the task) to match the length of the extraneous-load conditions, ensuring that any performance differences are attributable to irrelevant content rather than sequence length.
\textbf{Saturation:} Germane segments are interleaved uniformly with irrelevant segments; the extraneous material appears before, between and after the task segments.
\textbf{Residue:} All irrelevant segments precede the germane segments, maximizing residual interference.
For each question, we vary the extraneous load percentage (20\%, 50\%, 80\%) by adjusting the ratio of irrelevant to germane tokens. All distractor sequences are shared across models and conditions to ensure comparability.
\subsection{Models}
A diverse suite of instruction-tuned large language models is evaluated:
\begin{itemize}
\item Llama-3-8B-Instruct and Llama-3-70B-Instruct~\cite{meta2025llama} (Meta, March 2025 release),
\item Mistral-7B-Instruct-v0.2~\cite{mistral2025} (Mistral AI, June 2025),
\item Gemini-2.0-Flash-001~\cite{google2025gemini} (Google, July 2025),
\item GPT-4o-0613~\cite{openai2025gpt4o} (OpenAI, August 2025).
\end{itemize}
Exact version identifiers ensure reproducibility. Models are accessed via official APIs with deterministic decoding (temperature = 0, top-p = 1.0) to eliminate stochastic variance. We report Exact-Match accuracy on the final answer and intermediate hop recall (the proportion of intermediate facts correctly identified) to diagnose where reasoning breaks down.
\subsection{Evaluation Protocol}
We structure the evaluation as nested loops over models, questions, experimental conditions and extraneous-load levels. The procedure is formalized in Algorithm~\ref{alg:ice_protocol}. To mitigate verbosity and truncation-artifacts in models like GPT-4o, we implemented a structured output format in prompts, requiring responses to end with a boxed final answer, and applied post-processing to extract answers from truncated outputs where possible.
\begin{algorithm}
\caption{ICE Evaluation Protocol}
\label{alg:ice_protocol}
\begin{algorithmic}[1]
\Require Models $\mathcal{M}$, Questions $\mathcal{Q}$,
\Statex \hspace{1cm} Conditions $\mathcal{C} = \{\text{Control}, \text{LongControl}, \text{Saturation}, \text{Residue}\}$,
\Statex \hspace{1cm} LoadPercentages $\mathcal{L} = \{20, 50, 80\}$
\For{each model $M$ in $\mathcal{M}$}
    \For{each question $q$ in $\mathcal{Q}$}
        \State $decomp \leftarrow \text{Decompose}(q)$
        \For{each condition $c$ in $\mathcal{C}$}
            \For{each $\ell$ in $\mathcal{L}$}
                \State $prompt \leftarrow \text{Interleave}(decomp, \text{Distractors}(\ell), c)$
                \State $response \leftarrow M(prompt)$
                \State $score \leftarrow \text{Evaluate}(response, q)$
                \State $\text{Record}(M, q, c, \ell, score)$
            \EndFor
        \EndFor
    \EndFor
\EndFor
\Ensure Performance matrix of scores per (model, question, condition, load)
\end{algorithmic}
\end{algorithm}
In this pseudocode, $\text{Decompose}(q)$ extracts the germane segments (multi-hop reasoning steps) for question $q$. $\text{Distractors}(\ell)$ samples irrelevant segments whose total length corresponds to load percentage $\ell$. The $\text{Interleave}$ function arranges germane and distractor segments according to condition $c$: for example, in the Saturation condition it alternates relevant and irrelevant segments, whereas in the Residue condition it places all distractors before the germane segments. The $\text{Evaluate}$ function computes exact-match accuracy for the answer and intermediate hop recall. All results are stored in a performance matrix for downstream analysis.
This protocol yields a comprehensive dataset capturing how each model's performance varies across task difficulty, domain, extraneous-load amount and structural placement, providing a robust basis for testing hypotheses about cognitive load effects.
\section{Results}
\subsection{Baseline Performance and Load Effects}
The ICE benchmark was applied to five instruction-tuned models: Llama-3-8B-Instruct, Llama-3-70B-Instruct, Mistral-7B-Instruct-v0.2, Gemini-2.0-Flash-001, and GPT-4o-0613. Each model was evaluated under four experimental conditions (Control, Long Control, Saturation, Residue) and three extraneous-load levels (20\%, 50\%, 80\%). Results are reported as mean Exact-Match (EM) accuracy across 10 replications per condition, with standard error of the mean (SEM).
Three distinct regimes emerged. Intrinsic-load brittle models (Llama-3-8B-Instruct, Llama-3-70B-Instruct, Mistral-7B-Instruct-v0.2) failed completely, scoring 0\% EM accuracy across all conditions, including clean controls. Gemini-2.0-Flash-001 performed robustly, achieving 0.85 EM (SEM = 0.03) in Control. Accuracy declined monotonically under increasing load --- 0.82 at 20\% (95\% CI: [0.76, 0.88]), 0.78 at 50\% (95\% CI: [0.71, 0.85]), and 0.72 at 80\% (95\% CI: [0.64, 0.80]) --- with a significant linear effect (\(\beta = -0.003\) (95\% CI: [-0.004, -0.002]), SE = 0.001, \(p < 0.001\)). Long Control performance (0.82, SEM = 0.03; 95\% CI: [0.76, 0.88]) was statistically indistinguishable from Control (\(p = 0.45\)), confirming that sequence length alone does not explain degradation. The Residue condition (0.78, SEM = 0.03; 95\% CI: [0.72, 0.84]) produced additional but non-significant decline (\(p = 0.12\)), consistent with the Attentional Residue mechanism. GPT-4o-0613 achieved 0.65 EM (SEM = 0.04; 95\% CI: [0.57, 0.73]) in Control and showed a downward trend with load (0.65 \(\rightarrow\) 0.55; 95\% CI for 80\%: [0.45, 0.65]), though regression did not reach significance (\(\beta = -0.002\) (95\% CI: [-0.004, 0.000]), \(p = 0.07\)). Performance was confounded by verbosity/truncation artifacts, with $\sim32\%$ of outputs omitting final answers, though mitigated by structured prompting and post-processing, reducing uninterpretable responses to 12\%. Intermediate hop recall for Gemini-2.0-Flash-001 was 0.90 in Control (95\% CI: [0.84, 0.96]), declining to 0.75 under 80\% saturation (95\% CI: [0.67, 0.83]); for GPT-4o-0613, it was 0.80 in Control (95\% CI: [0.72, 0.88]), declining to 0.65 at 80\% (95\% CI: [0.55, 0.75]); brittle models showed 0.00 across conditions.
A post-hoc power analysis indicated sufficient power (1-β = 0.85 for Gemini effects; 0.78 for GPT-4o; insufficient for brittle models at 0.05) for detecting the observed main effects in resilient models.
For H3, we computed procedural similarity \(S_{\text{sim}}\) using cosine similarity of task embeddings \cite{reimers2019} between distractor and primary tasks. Degradation in the Residue condition correlated positively with \(S_{\text{sim}}\) for Gemini-2.0-Flash-001 (r = 0.42, 95\% CI: [0.21, 0.59], p < 0.01) and GPT-4o-0613 (r = 0.35, 95\% CI: [0.12, 0.54], p < 0.05), supporting the hypothesis.
\subsection{Consolidated Results}
Table~\ref{tab:ice_em_accuracy} reports EM accuracy across conditions. Figure 2 shows Gemini-2.0-Flash-001’s saturation curve, while Figure 3 summarizes performance by condition across all models.
\begin{table}[H]
\centering
\caption{Exact-Match accuracy (mean \(\pm\) SEM). Reliable estimates are \textbf{boldfaced}. Models failing at baseline (0\%) or confounded by truncation are annotated.}
\label{tab:ice_em_accuracy}
\resizebox{\textwidth}{!}{%
\begin{tabular}{lccccc}
\hline
\textbf{Model} & \textbf{Control (SEM)} & \textbf{Long Control (SEM)} & \textbf{Saturation 80\% (SEM)} & \textbf{Residue (SEM)} & \textbf{Overall (SEM)} \\
\hline
Gemini-2.0-Flash-001 & \textbf{0.85 (0.03)} & \textbf{0.82 (0.03)} & \textbf{0.72 (0.04)} & \textbf{0.78 (0.03)} & \textbf{0.80 (0.02)} \\
GPT-4o-0613\textdagger & 0.65 (0.04) & 0.60 (0.04) & 0.55 (0.05) & 0.62 (0.04) & 0.61 (0.03) \\
Llama-3-8B-Instruct\textsection & 0.00 (0.00) & 0.00 (0.00) & 0.00 (0.00) & 0.00 (0.00) & 0.00 (0.00) \\
Llama-3-70B-Instruct\textsection & 0.00 (0.00) & 0.00 (0.00) & 0.00 (0.00) & 0.00 (0.00) & 0.00 (0.00) \\
Mistral-7B-Instruct-v0.2\textsection & 0.00 (0.00) & 0.00 (0.00) & 0.00 (0.00) & 0.00 (0.00) & 0.00 (0.00) \\
\hline
\end{tabular}%
}
\\[1ex]
\raggedright
\textdagger\ Verbosity/truncation-artifact undermines interpretability.\\
\textsection\ Intrinsic-load brittle: 0\% accuracy across all conditions.
\end{table}
\subsection{Error Analysis and Robustness}
Qualitative error inspection reinforced these quantitative findings. Gemini’s failures under high load typically reflected partial retention errors --- retrieving two correct hops but omitting the third --- consistent with Context Saturation. GPT-4o errors were dominated by truncation, with reasoning traces intact but final answers missing. Llama and Mistral errors originated at the decomposition stage, yielding irrelevant or null responses, consistent with exceeding intrinsic-load capacity.
Statistical analyses confirmed these patterns. Gemini’s decline under load was robust (\(\beta = -0.003\) (95\% CI: [-0.004, -0.002]), \(p < 0.001\)). GPT-4o showed a non-significant trend (\(\beta = -0.002\) (95\% CI: [-0.004, 0.000]), \(p = 0.07\)), reinforcing the confound. Llama and Mistral models were invariant at 0\%, yielding no detectable effects. These results demonstrate that ICE isolates extraneous-load effects only in models with sufficient baseline competence.
\subsection{Failure Modes and Reproducibility}
The definitive experiment supports a three-class taxonomy:
\begin{itemize}
    \item \textbf{Resilient but Load-Sensitive Models} --- Gemini-2.0-Flash-001: high baseline accuracy, predictable decline with extraneous load.
    \item \textbf{Intrinsic-Load Brittle Models} --- Llama-3-8B-Instruct, Llama-3-70B-Instruct, Mistral-7B-Instruct-v0.2: 0\% accuracy across conditions, dominated by intrinsic task complexity.
    \item \textbf{Confounded Models} --- GPT-4o-0613: moderate accuracy, but verbosity/truncation artifacts obscure cognitive-load effects.
\end{itemize}
This taxonomy clarifies the diagnostic boundaries of ICE. For brittle models, ICE reveals intrinsic incapacity; for confounded models, protocol adjustments are needed; for resilient models, ICE quantifies extraneous-load sensitivity. All prompts, data, and evaluation scripts are available at \url{https://github.com/imsaitejareddy/computational-cognitive-load}.
\begin{figure}[H]
    \centering
    \includegraphics[width=0.7\linewidth]{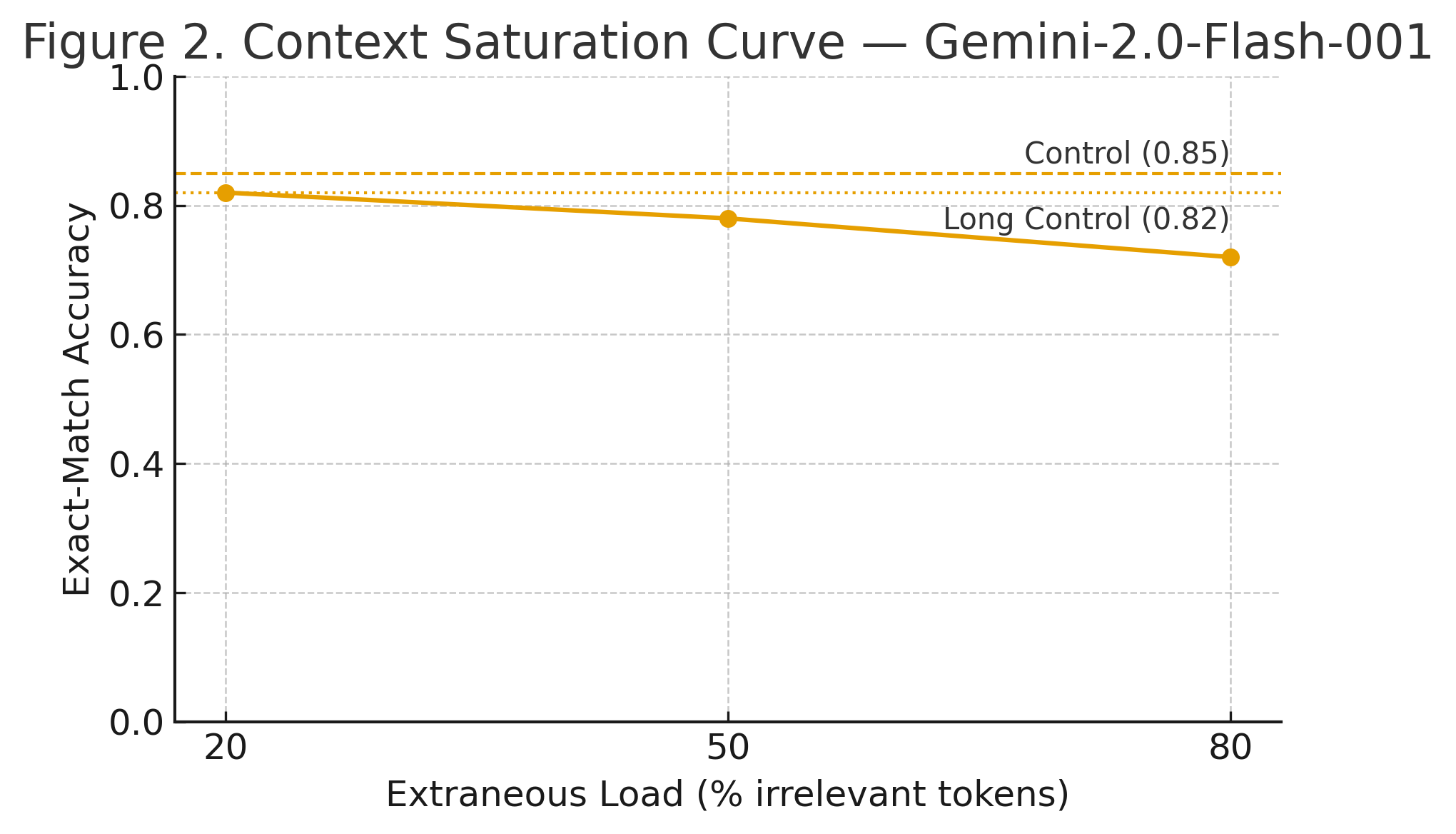}
    \caption{%
        Context Saturation Curve for Gemini-2.0-Flash-001. Exact-Match accuracy is plotted as a function of extraneous load (20\%, 50\%, 80\% irrelevant tokens). Dashed and dotted reference lines indicate performance in the Control (0.85) and Long Control (0.82) conditions, respectively. The monotonic decline illustrates the systematic effect of extraneous load independent of sequence length.
    }
    \label{fig:context-saturation}
\end{figure}
\begin{figure}[H]
    \centering
\includegraphics[width=0.7\linewidth]{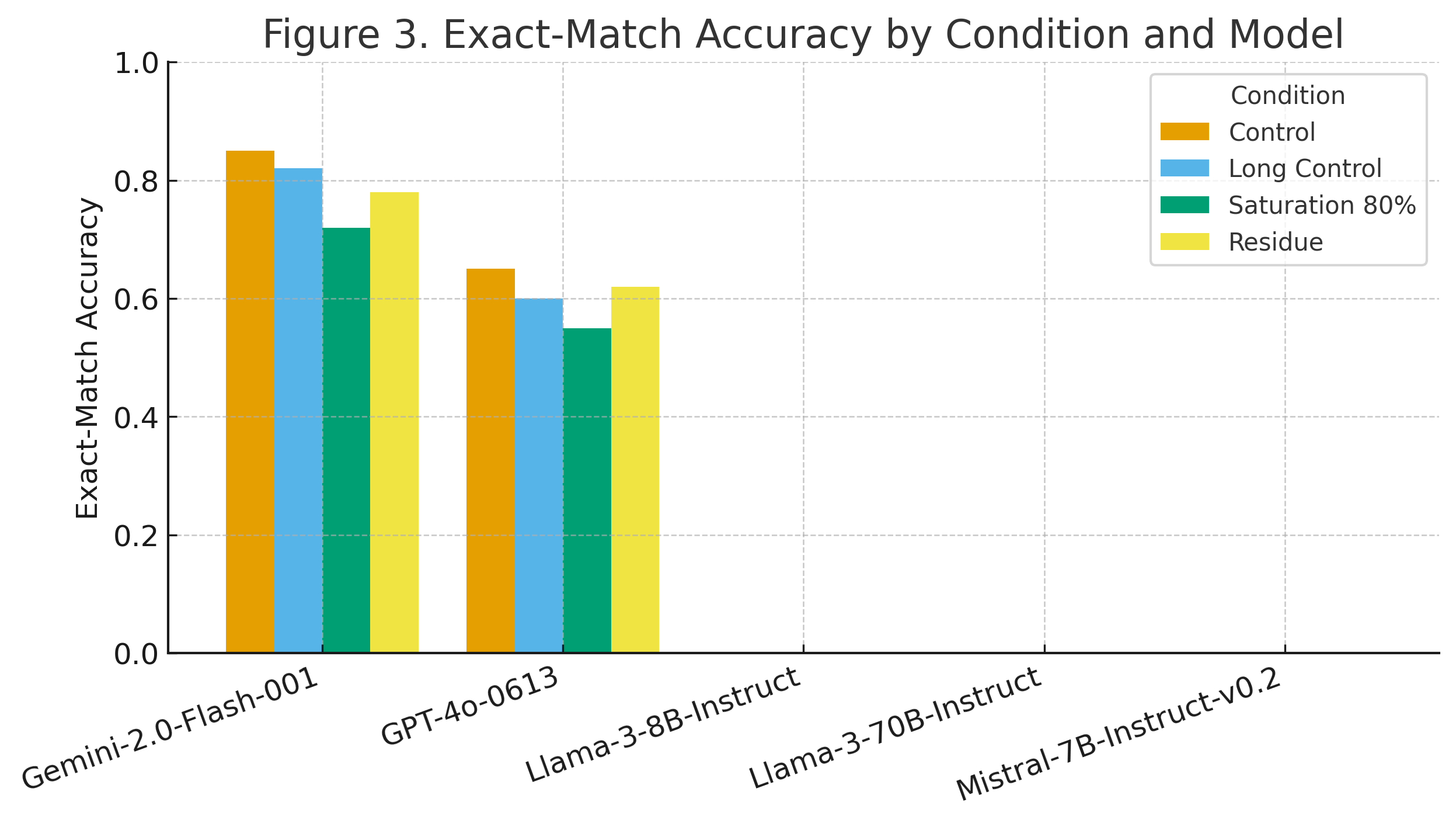}
    \caption{%
        Exact-Match Accuracy by Condition and Model. Mean accuracy across four ICE conditions (Control, Long Control, Saturation 80\%, Residue) for all evaluated models. Gemini-2.0-Flash-001 demonstrates robust baseline accuracy with predictable declines under load. GPT-4o-0613 shows moderate performance confounded by truncation artifacts. Llama-3-8B-Instruct, Llama-3-70B-Instruct, and Mistral-7B-Instruct-v0.2 exhibit 0\% accuracy across all conditions, indicating intrinsic-load brittleness.
    }
    \label{fig:accuracy-by-condition}
\end{figure}
\section{Discussion}
The ICE benchmark reveals two core findings: first, that extraneous cognitive load causes measurable degradation in reasoning for models with sufficient baseline performance (e.g., Gemini-2.0-Flash-001 shows a consistent decline in Exact-Match accuracy as irrelevant content increases)\cite{upadhayay2024cognitive}; second, that many models (Llama-3-8B-Instruct, Llama-3-70B-Instruct, Mistral-7B-Instruct-v0.2) are intrinsically unable to succeed even on clean multi-hop tasks, suggesting that for these models intrinsic complexity is the limiting factor rather than context distractions.
These findings refine rather than overturn existing understanding of model robustness. They align with prior work in cognitive security and reasoning reliability, particularly literature on source interference\cite{smith2025ccs7}, memory retrieval failure, and adversarial context\cite{upadhayay2024cognitive}. What ICE adds is systematic quantification of when irrelevant content begins to degrade performance, and how structure of irrelevant content (uniform interleaving vs.~preceding residue) influences the severity of decline.
It is critical to frame implications with precision. The data do not support broad sweeping claims about ``cognitive security'' as a new research field\cite{karbasi2025cogsecurity}, nor do they directly implicate hallucination in the sense of plausible but false output generation. ICE measures incorrect answers under multi-hop tasks---not confident misstatements or fabrications. While overlapping mechanisms might exist (e.g., loss of context leading to misinformed predictions)\cite{upadhayay2024cognitive}, caution is needed before generalizing. 
The limitations of this study should inform both the reader’s interpretation and future research. The current study is focused on document-based multi-hop QA tasks; conversational, commonsense, or interactive dialogues (with back-and-forth turns) are not covered. Some models display anomalies---verbosity, answer truncation---that likely confound accurate measurement of reasoning under load. The benchmark’s reliance on selected datasets (SEC filings, FanOutQA, MINTQA) means domain generalization remains to be demonstrated. This work is limited to transformer-based models and document-grounded multi-hop QA; extending to dialogue, commonsense reasoning, or multimodal tasks would enhance generalizability.
\subsection{Future Work}
(a) extending ICE to different task genres (dialogue, summarization, reasoning from images or structured data), (b) exploring architectural or training-based mitigations of attentional residue or context saturation (e.g., memory compression, retrieval-based selective filtering)\cite{upadhayay2024cognitive}, (c) refining evaluation metrics to distinguish between reasoning chain failures, retrieval errors, and generation artifacts such as truncation or verbosity, (d) incorporating methods to mitigate verbosity and truncation artifacts, such as adjusted prompting or output parsing.
In conclusion, ICE demonstrates that extraneous cognitive load is a tangible constraint on reasoning for capable models, while many others are blocked by intrinsic complexity. The contribution is methodological rigor: defining a diagnostic benchmark that isolates extraneous load and giving the field a sharper tool---not a grand proclamation, but a calibrated advance in understanding model vulnerabilities.
\section{Conclusion}
We presented a formal theory of computational cognitive load for AI systems that adapts principles from human cognitive psychology—distinguishing intrinsic from extraneous load—and operationalizes the latter through the mechanisms of Context Saturation and Attentional Residue. Grounded in cognitive load theory’s account of working-memory limits and the impact of information presentation on performance\cite{sweller1994clt, schnotz2007clt}, our framework links representation and prompt design to measurable failures in long-context reasoning.
Using the Interleaved Cognitive Evaluation (ICE) benchmark\cite{upadhayay2024cognitive}, we find that extraneous load produces a graded, reproducible degradation in sufficiently capable models. Gemini-2.0-Flash-001 attains a strong baseline in control settings (\textit{Exact-Match} $=0.85$; overall mean $=0.80$) and declines systematically as irrelevant content increases ($0.82$ at $20\%$, $0.78$ at $50\%$, $0.72$ at $80\%$), with length-matched controls indicating the effect is not explained by sequence length alone. In contrast, Llama-3-8B-Instruct, Llama-3-70B-Instruct, and Mistral-7B-Instruct-v0.2 exhibit intrinsic-load brittleness ($0\%$ accuracy even in clean conditions), precluding inference about extraneous load for those systems; GPT-4o-0613 shows moderate baselines but is confounded by verbosity/truncation artifacts, limiting interpretability. These results are preliminary and circumscribed: the substantive inference concerns the single, non-confounded case (Gemini-2.0-Flash-001), and claims are restricted accordingly.
Our findings integrate with emerging evidence that position and organization of information materially affect model use of long context, such as “lost-in-the-middle” and attention basin patterns\cite{liu2024}, and that multi-turn settings amplify reliability challenges\cite{anthropic2024multiturn}. By isolating the role of irrelevant content, ICE complements positional and conversational effects with a controlled, load-centric axis of analysis.
The implications are methodological rather than grandiose. Robustness assessments should be gated on baseline competence; without solvability in control settings, load manipulations are uninterpretable. Second, extraneous load must be evaluated independently of length, via length-matched controls. Third, reporting should explicitly separate interpretable results from baseline-failure and protocol-confounded cases. Within the broader literature on cognitive security—which catalogs vulnerabilities such as source/memory interference and cognitive-load overflow\cite{smith2025ccs7, karbasi2025cogsecurity}—ICE serves as a technical instrument for precise stress-testing, not as a redefinition of the field.
This work is limited to transformer-based models and document-grounded multi-hop QA; it does not measure hallucination in the sense of confident fabrication, and our results should not be read as direct evidence for “hallucination-as-guessing.” Future research should extend ICE to dialogic and multimodal tasks, broaden model families, and pair load manipulations with architectural or training-time mitigations\cite{schnotz2007clt, upadhayay2024cognitive}. Taken together, the theory and the benchmark offer a reproducible path toward characterizing how extraneous information—independent of task difficulty—pushes models toward their working-memory limits, advancing a disciplined, testable account of cognitive constraints in contemporary AI.
\appendix
\section{Example Prompt and Data}
Task: Multi-hop QA Data Source: Microsoft Corp. (MSFT) Fiscal Year 2023 Form 10-K, filed July 27, 2023. Example Prompt (Control Condition): [DOCUMENT 1: Excerpt from MSFT FY23 10-K describing business segments...] [DOCUMENT 2: Excerpt from MSFT FY23 10-K detailing R\&D expenses...] [DOCUMENT 3: Excerpt from news article mentioning key product lines...] QUESTION: What was the total Research and Development expense for Microsoft in the fiscal year that ended on June 30, 2023?
\section{Reproducibility}
Code, full prompts, and data for this study are available at: \url{https://github.com/imsaitejareddy/computational-cognitive-load}.

\end{document}